\documentclass[a4paper]{svproc}
\usepackage{graphicx} 
\usepackage{wrapfig}
\usepackage{multicol}
\usepackage{footmisc}
\usepackage{cite}
\usepackage{graphicx}
\usepackage{multirow}
\usepackage{type1cm}

\makeatletter
\renewcommand\paragraph{\@startsection{paragraph}{4}{\z@}%
    {3.25ex \@plus1ex \@minus.2ex}%
    {-1em}%
    {\normalfont\normalsize\bfseries}}
\makeatother

\begin{document}
\mainmatter              
\title{Robotic Fabric Flattening with Wrinkle Direction Detection}
\titlerunning{Robotic Fabric Flattening with Wrinkle Direction Detection} 
\author{Yulei Qiu\inst{1}, Jihong Zhu\inst{1,2}, Cosimo Della Santina\inst{1}, Michael Gienger\inst{3} and Jens Kober\inst{1}}
\authorrunning{Yulei Qiu et al.} 
\tocauthor{Yulei Qiu, Jihong Zhu, Cosimo Della Santina, Michael Gienger and Jens Kober} 
\institute{
Delft University of Technology, the Netherlands,\\
\email{yulei.qiu@outlook.com,\{j.zhu-3, c.dellasantina, j.kober\}@tudelft.nl}\\
\and
University of York, the UK,\\
\email{jihong.zhu@york.ac.uk}\\
\and
Honda Research Institute Europe, Germany,\\
\email{michael.gienger@honda-ri.de}\\
} 

\maketitle


\begin{abstract}
Deformable Object Manipulation (DOM) is an important field of research as it contributes to practical tasks such as automatic cloth handling, cable routing, surgical operation, etc. Perception is considered one of the major challenges in DOM due to the complex dynamics and high degree of freedom of deformable objects. In this paper, we develop a novel image-processing algorithm based on Gabor filters to extract useful features from cloth, and based on this, devise a strategy for cloth flattening tasks. We also evaluate the overall framework experimentally and compare it with three human operators. The results show that our algorithm can determine the direction of wrinkles on the cloth accurately in simulation as well as in real robot experiments. Furthermore, our dewrinkling strategy compares favorably to baseline methods. The experiment video is available on https://sites.google.com/view/robotic-fabric-flattening/home
\end{abstract}

\section{Introduction}
Despite significant progress made in recent years, DOM is still considered one of the major challenges in robotics due to the complex dynamics and high degree of freedom of deformable objects \cite{Zhu2022}. Robots working in a human environment are expected to have good capability to handle them, as many manipulation tasks in daily life involve deformable objects, such as picking up fruits and folding clothes. Like many other work, DOM tasks can be broken down into perception and control. Perception in DOM is to obtain states to describe the object, while control uses this information to guide robot motion. The goal of perception and control in DOM is to efficiently infer the complex configuration of a deformable object, to determine a policy for manipulation and to perform a corresponding action.


Therefore, this paper concerns itself with perception and control in cloth manipulation, which has attracted much attention over the years \cite{Sun2016}. Early perception methods, for example, rely on extra markers \cite{Bersch2011} that require the cloth to be completely covered to infer the state, or predefined visual features \cite{Miller2011, Stria2014} which approximate the cloth with predefined polygon models. The use of markers is usually not possible in practical cases, and predefined geometric features are often not robust and introduce errors in state estimation \cite{Ma2022}. Actions, policies, or strategies to manipulate can be learned from human demonstrations since human behavior during the cloth manipulation process is difficult to model. Commonly used methods in Learning from Demonstration (LfD) include Dynamic Movement Primitives (DMPs) \cite{Saveriano2021} and Gaussian Mixture Models (GMMs) \cite{calinon2016tutorial}. Recent data-driven methods combine perception and control stages, which do not infer the explicit states of the cloth. Rather, they output the policy directly, which learn the mapping from the raw RGB(D) images to the robot actions \cite{Wu2019, Jangir2019, Tsurumine2019}. Like most learning-based methods, they require a large amount of training data and are computationally demanding.

In this paper, we proposed a perception algorithm that does not rely on markers or geometry shapes, followed by a control strategy to make use of the sensing information to flatten fabric. Our method calculates the magnitude and direction of wrinkles on the cloth, thereby outputs a 2D vector to represent the stretching direction which the robot can drag along to remove wrinkles, and an operation point where the robot touches the fabric. The magnitude reflects the distribution of wrinkles, which helps in finding the operation point. The outcomes from the perception can be directly utilized by the the robot manipulator to execute the flattening tasks.

The main contributions of this paper are:
\begin{enumerate}
    \item A novel image-processing algorithm on cloth-like deformable objects that computes a stretching direction and an operation point which can be directly used by the robot manipulator.
    \item A simple control strategy leveraging the output of the perception step to perform the flattening actions.
    \item A framework combining the algorithm and control strategy that enables us to perform cloth flattening tasks in high efficiency compared to baseline methods.
\end{enumerate}
%
\vspace{-3mm}


%
\section{Related Work}
The following literature shows some related work in the perception and control about cloth manipulation.

\textbf{Perception:} Besides aforementioned perception methods \cite{Bersch2011, Miller2011, Stria2014}, other work on cloth perception uses wrinkle as an intuitive visual feature \cite{Ramisa2012, Caporali2020}, since they are obvious when clothes are placed on a flat surfaces. These two approaches calculate a heat map from visual data, which determines the states or the grasping policies. Prior to them, a Gabor filter \cite{Lee1996} is applied on the images to extract wrinkle features \cite{Yamazaki2009}. The Gabor filter with designed kernel can effectively detect the area of the cloth exhibiting wrinkles. Jia et al. \cite{Jia2017} proposed HOW features corresponding to shadow variation computed by applying Gabor filters with multiple orientations and wavelengths. We notice from the previous work that visual descriptors enable extracting low-dimensional features of cloth-like objects from RGB-(D) data, thereby forming an effective state representation, which holds potential for utilization in other fabric manipulation tasks.

\textbf{Control:} Robot actions can be either determined by a controller based on the perception information (model-based) \cite{Yan2020, Chang2020, Hoque2020} or learned from the raw image input (model-free) \cite{Matas2018, Singh2019}. Model-based methods are generally more computationally efficient and can provide a better understanding of the underlying physics of the system, but require more prior knowledge and may not be as flexible as model-free control \cite{Zhang2021, Pinosky2022}. Model-free methods like Imitation Learning (IL), demonstrate superior performance in the learning of movement policies, as they are capable of capturing the trajectory that predicts the actions of the expert given the corresponding RGB-D images \cite{Joshi2017, Jia2017, Salhotra2022}. These methods are more flexible and can be used when the dynamics of the system are not well understood, yet they require more data and can be more computationally intensive \cite{Zhang2021, Pinosky2022}, which is not practical in our applications.
\vspace{-3mm}

\begin{figure}[tb]
    \centering
    \includegraphics[width=0.8\textwidth]{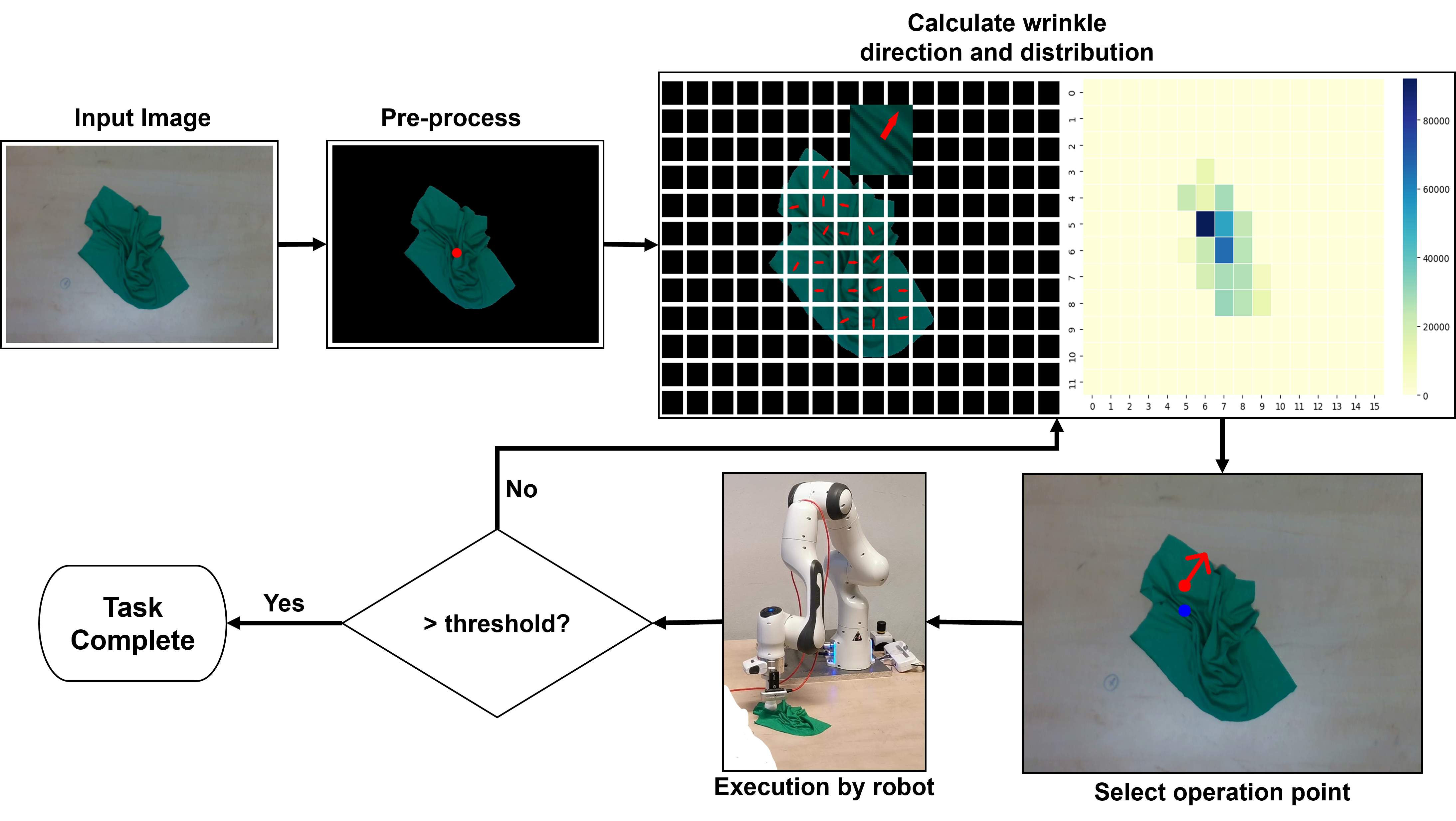}
    \caption{Pipeline of the proposed framework. \textit{Pre-process:} The RGB image from the camera is first processed by HSV thresholding to remove the background, and the \textbf{Center of Mass} (\textbf{CoM}, the red dot) is calculated. \textit{Calculate wrinkle direction and distribution:} The image is split into several small blocks and Gabor filters are applied to them. Then for each block we calculate a direction in which the robot should move to effectively remove the wrinkle. \textit{Selection operation point:} The operation point is determined according to the distribution of wrinkles. \textit{Execution by robot:} The robot starts to remove the wrinkles.}
    \label{fig:outline}
\end{figure}

\section{Technical Approach}


Given an input RGB image from a top-down view of a cloth-like object, the algorithm can accurately identify the wrinkle direction and distribution on the cloth, thereby computing both the stretching direction and operation point. The stretching direction is perpendicular to the wrinkle, informing the direction along which the robot should apply force to remove wrinkles. The magnitude of the wrinkle indicating the distribution of the wrinkles helps determine the operation point, which is the location where the robot touches the fabric then moves along the stretching direction. Once the direction and operation point have been determined, the robot can execute the actions to flatten the cloth. After one step of action, the robot moves up again to get a global image and runs the algorithm again. The pipeline as shown in Fig.~\ref{fig:outline} will work iteratively until the cloth is flattened.


\subsection{Calculate Stretching Direction}

Gabor filters are effective in extracting wrinkles in images, hence are useful to infer the directions and distribution of wrinkles in an image. A 2D Gabor filter can be mathematically represented as:
\begin{equation}
    g(x,y;\lambda,\phi,\sigma,\gamma,\theta) = \exp{\left(-\frac{x'^{2}+\gamma^{2}y'^{2}}{2\sigma^{2}}\right)} \exp{\left(j(2\pi\frac{x'}{\lambda}+\phi)\right)}
\end{equation}
where $x'=x\cos{\theta}+y\sin{\theta}$, $y'=-x\sin{\theta}+y\cos{\theta}$, $\lambda$ is the wavelength, $\phi$ is the phase offset, $\sigma$ is the standard deviation of the Gaussian, $\gamma$ is the spatial aspect ratio, and $\theta$ is the orientation of the normal to the parallel stripes of the Gabor filter.

As shown in Fig.~\ref{fig:outline}, the raw RGB image will first be processed to remove the background. Then the \textbf{Center of Mass} (\textbf{CoM}, denoted by red dot) of the cloth is computed. The image $I$ is split into $N_{b}$ small blocks, $I_{j}$ ($j=0,1,...N_{b}-1$). 
For each block $I_{j}$, Gabor filters in the following orientations are applied: $\theta_{i} = i \times \frac{\pi}{N_{o}}, \forall i\in\{0,1,...,N_{o}-1\}$
where $N_{o}$ is the number of orientations.

\begin{wrapfigure}{R}{0.35\textwidth}
    \vspace{-.5cm}
    \centering
    \includegraphics[width=0.35\textwidth]{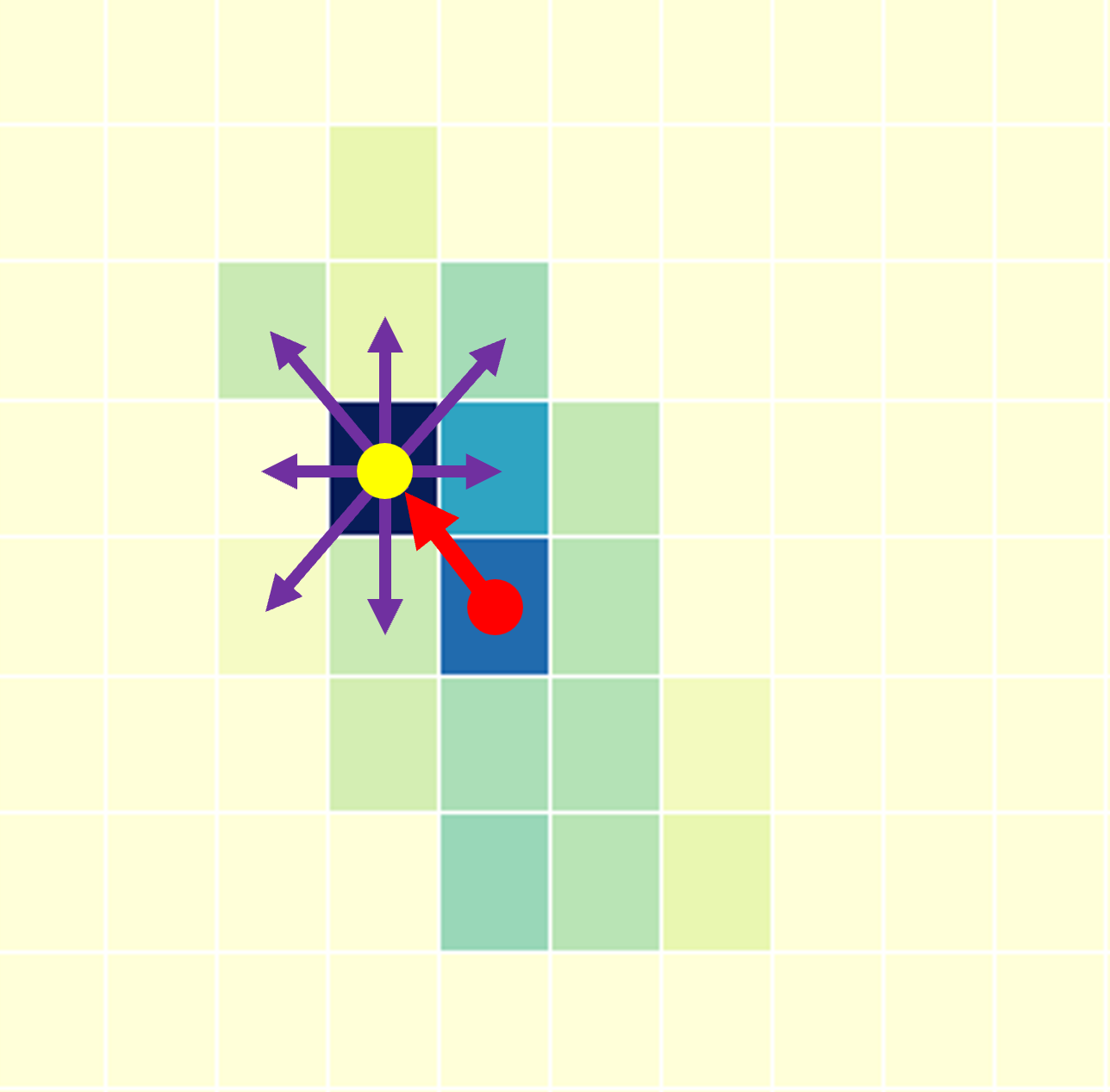}
    \caption{Illustration of operation point selection. The yellow dot is the block with highest magnitude. The red arrow points at it from the CoM. The purple arrows end at the candidate operation points.}
    \label{fig:point-select}
    \vspace{-1cm}
\end{wrapfigure}

The magnitude $M_{w,(i,j)}$ of wrinkles of the $j$-th block $I_{j}$ in a particular orientation $\theta_{i}$ is quantified by computing the sum of values across all pixels in the corresponding output $I_{\mathrm{ori},(i,j)}$ of the Gabor filter in that orientation 
$M_{w,(i,j)} = \sum_{p\in I_{\mathrm{ori},(i,j)}} {p}.$
Note that $I_{\mathrm{ori},(i,j)}$ is an intensity map of image $I_{j}$ in the given orientation $\theta_{i}$, and $p$ corresponds to the value at pixel $p$ in $I_{\mathrm{ori},(i,j)}$. Therefore, the magnitude $M_{w,j}$ of wrinkles in $I_{j}$ is determined by the highest value among the all orientations: 
$M_{w,j} = \max{M_{w,(i,j)}},\ \forall i \in \{0,1,...,N_{o}-1\}$

\subsection{Select Operation Point}
Since the magnitude of a block reflects how wrinkled it is, we first find the block with highest magnitude, which is consider as the most wrinkled block. However, rather than directly operating on this block, the operation point will be selected among the neighboring blocks to avoid touching the existing wrinkle too much and create new wrinkles. We measure the angle between red and purple arrows (see Fig.~\ref{fig:point-select}), and choose those with angle less than 90 degrees to make sure they are point outwards. The finally-determined point is randomly select from the blocks that bring an outward direction.

\subsection{Execution}
After the operation point is selected, the robot will execute the motion accordingly. The end-effector first lower to touch the operation point, like a finger pressing the cloth on the table, and the manipulator will move along the stretching direction. The moving distance is fixed.

\section{Experiments}
In order to evaluate our proposed method, we conducted a series of experiments in both simulated and real-world scenarios. Specifically, we first validated our approach on wrinkled cloth within the SoftGym simulation environment \cite{Lin2020}. Additionally, we conducted real-world experiments using a Franka Emika Robot.
\subsection{Simulation}
\begin{figure}[t]
    \centering
    \includegraphics[width=1\textwidth]{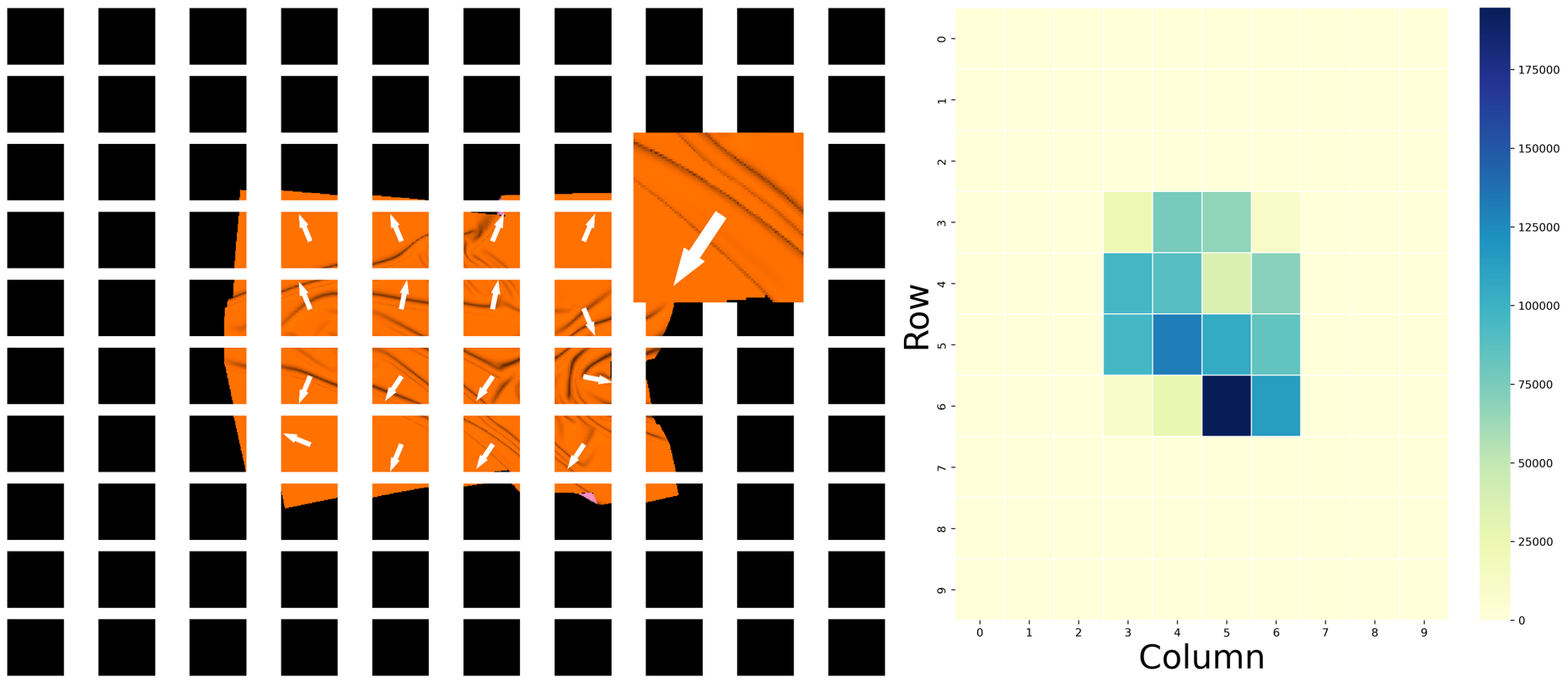}
    \caption{Perception result in simulation. \textbf{Left}: Calculated Stretching Direction. \textbf{Right}: Heatmap of Magnitude.}
    \label{fig:sim}
\end{figure}

We employed the perception algorithm in Softgym \cite{Lin2020}. In this simulation environment, we were able to rapid-prototype and fine-tune the hyperparameters of the Gabor filter. The observation is an RGB image of $720 \times 720$ pixels captured from a top-down perspective. As shown in the left part of Fig.~\ref{fig:sim}, the calculated  directions are perpendicular to their corresponding wrinkles. Besides, the heatmap in the right part of the figure displays the distribution of the wrinkles, where higher magnitude indicates more wrinkles. We can see that the distribution shown in heatmap is almost the same as that shown in the left part.
\subsection{Real Robot}
\begin{figure}[t]
    \centering
    \includegraphics[width=0.6\textwidth]{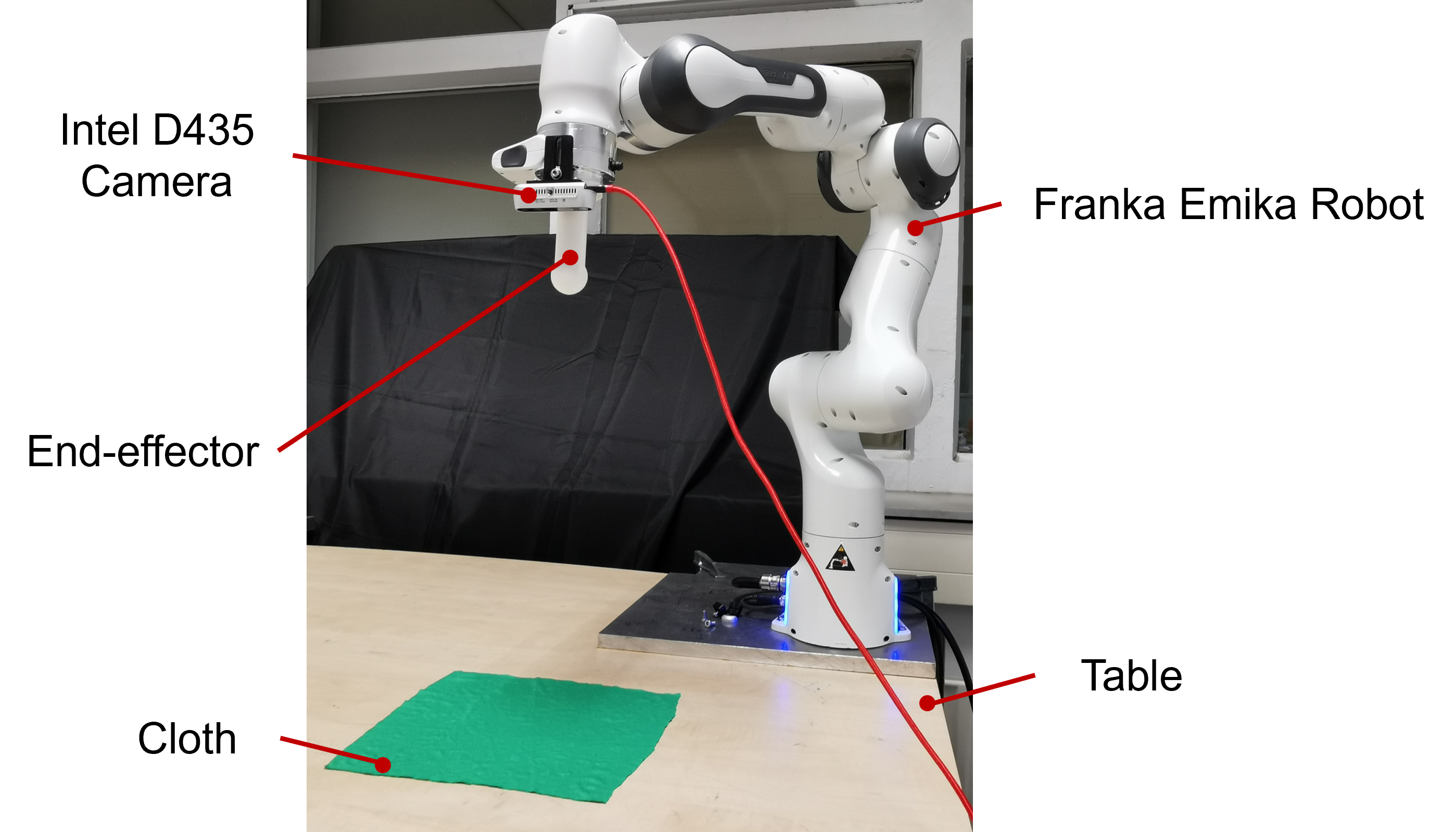}
    \caption{Experiment setup}
    \label{fig:exp-robot}
\end{figure}

To better make use of the direction and operation point in the cloth manipulation task, we designed an end-effector and manufactured it using 3D printing. Fig.~\ref{fig:exp-robot} depicts the end-effector mounted on the manipulator. The end-effector has a finger-like shape, which is a combination of a cylinder and a ball. During the manipulation, the ball part comes into contact with the cloth. Therefore, the process is restricted in a 2D surface. An impedance controller \cite{Franzese2021} is deployed on the manipulator to execute the motion, given the target coordinate relative to the robot.

We evaluate our proposed method on a Franka Emika Robot. In order to obtain RGB images of the cloth on the table from a top-down perspective as input, we mounted an Intel RealSense D435 camera onto the end-effector. Since the position of operation point calculated by the algorithm is in the pixel coordinate system, i.e., on the image, we need to transform it into robot spatial coordinates. The position in the world coordinate system $^\mathrm{base}P$ can be calculated by the following equation:
$^\mathrm{base}P = ^\mathrm{base}T_\mathrm{tool} \times ^\mathrm{tool}T_\mathrm{cam} \times ^\mathrm{cam}P
$,
where $^\mathrm{base}T_\mathrm{tool}$ is the transformation from end-effector (tool) to world (robot) coordinate system, $^\mathrm{tool}T_\mathrm{cam}$ is from camera relative to end-effector coordinate system, and $^\mathrm{cam}P$ is the position in camera coordinate system. $^\mathrm{base}T_\mathrm{tool}$ can be found by reading the pose of the end-effector from the robot sensor. $^\mathrm{tool}T_\mathrm{cam}$ is computed by hand–eye calibration calibration, for which we use OpenCV's chessboard \cite{opencv_library} and the Visual Servoing Platform (ViSP) \cite{Marchand2005} to calculate the transformation. $^\mathrm{cam}P$ can be calculated given the position in pixel coordinate system. The transformation from pixel coordinate to camera coordinate is determined by intrinsic parameters of the camera. The camera used in our experiment, Intel D435, provides an SDK for intrinsic transformation.

To conclude, we calculate the position of operation point in pixel coordinate. Through a series of transformation, it is converted to the position in robot coordinate so that the manipulator can execute the motion. 



\subsection{Baselines}
We conducted a comparative analysis of the performance of our algorithm against three other baseline methods: Random, Heuristic, and Human. The Random method randomly selects an operation point within the cloth, while the Heuristic method selects the operation point from candidate points obtained by applying a corner detection algorithm on the cloth. For both Random and Heuristic methods, the stretching direction is determined by the unit vector pointing from the CoM to the operation point. On the other hand, the Human method involves a human operator who manually selects the operation point and stretching direction. During this task, the participant observes the cloth from the image captured by the camera and determines the manipulation policy by clicking on the window displayed on the screen. The position of the first click corresponds to the operation point, while the stretching direction is determined by a unit vector pointing from the first click to the second click. We collected experimental data from three human operators. Table \ref{tab:baselines} summarizes how each method determines the manipulation policy.


\begin{table}[tb]
\caption{Manipulation policy of baseline methods.}
\label{tab:baselines}
\begin{center}
\resizebox{0.9\textwidth}{!}{%
\begin{tabular}{|c|c|c|}
\hline
Name & Operation Point & Direction \\ \hline
Random & Randomly selected within the cloth & From CoM to OP \\ \hline
Heuristic & Randomly selected around edges & From CoM to OP \\ \hline
Human & Selected by operator & Determined by operator \\ \hline
\end{tabular}%
}
\end{center}
\end{table}

\subsection{Evaluation metric}
The coverage is defined as the number of pixels in the cloth over the total number of pixels in the image, which is a measurement of area of the cloth. Then two types of coverage are used in the evaluation and as stopping criterion. The first one is the initial coverage, which is the coverage after generating some wrinkles for the first time. The second type is the relative coverage, which is defined as the ratio of the final coverage ($F$) to the coverage when the cloth is fully flattened ($C$):
$R = \frac{F}{C}.$
The stopping criterion is achieving a relative coverage greater than 99\%. Throughout the task, the relative coverage is calculated at each step. Additionally, we measure the number of steps required to meet the stopping criterion as another metric.
\subsection{Procedures}
The workflow of the experiment is:
\begin{enumerate}
    \item Place the cloth in the middle of the camera's field of view and manually flatten it before starting the experiment.
    \item Capture the image of cloth when it is fully flatten and calculate the full coverage.
    \item Manually make some wrinkles on the cloth.
    \item The algorithm estimates the coverage and computes the stretching direction and operation point.
    \item The robot manipulator touches the operation point with its end-effector and drags a fixed distance along the stretching direction.
    \item The robot moves back up to the starting position to to capture a global image, completing one step of the task.
    \item Repeat Step 4-6 until the coverage meets the stopping criterion.
\end{enumerate}

\section{Experimental Results and Discussion}
\begin{figure}[t]
    \centering
    \includegraphics[width=0.95\textwidth]{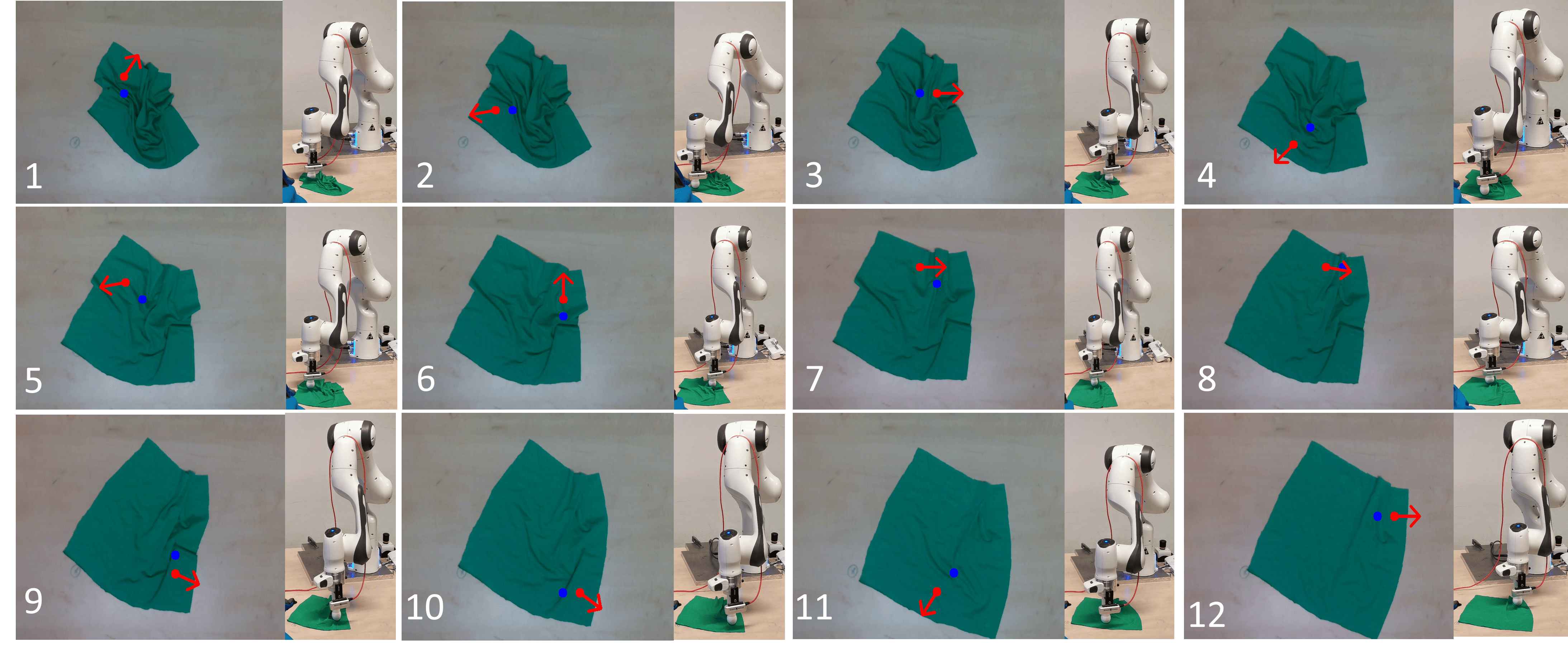}
    \caption{Cloth flattening process using Proposed method.}
    \label{fig:real-sequence}
\end{figure}
The experiment video is available on the website in the abstract. Fig.~\ref{fig:real-sequence} shows a continuous sequence of the robot performing the operation using the Proposed method. In the real robot experiment, the initial coverage of all tasks are clustered using k-means into two types, so they are categorized as either ``Easy'' or ``Hard'' based on their initial coverage. The cluster with a high initial coverage is considered as ``Easy'', while the other one with a low initial coverage is considered ``Hard''. Table~\ref{tab:exp} shows the number of easy and hard tasks for different methods, as well as the average number of steps required to complete these two types of tasks. In this experiment, there are three human operators and every participant performs 5 tasks for each method. We can see that Table~\ref{tab:exp} shows the ratio of different difficulty is approximately easy:hard=1:2. Given a similar proportion of easy and hard tasks for each method, comparing the average number of steps required to complete the tasks is a fair way to evaluate the efficiency. Overall, the Human method outperforms all other methods for both easy and hard tasks, completing tasks in $4.50$ and $6.45$ steps, respectively. For easy tasks, the Heuristic method performs better than the Random and Proposed methods, requiring only $5.80$ steps compared to $7.00$ steps for the other two methods. However, for hard tasks, the Heuristic and Random methods require more steps than the Proposed method to complete the task.

\begin{figure}[t]
    \centering
    \includegraphics[width=0.6\textwidth]{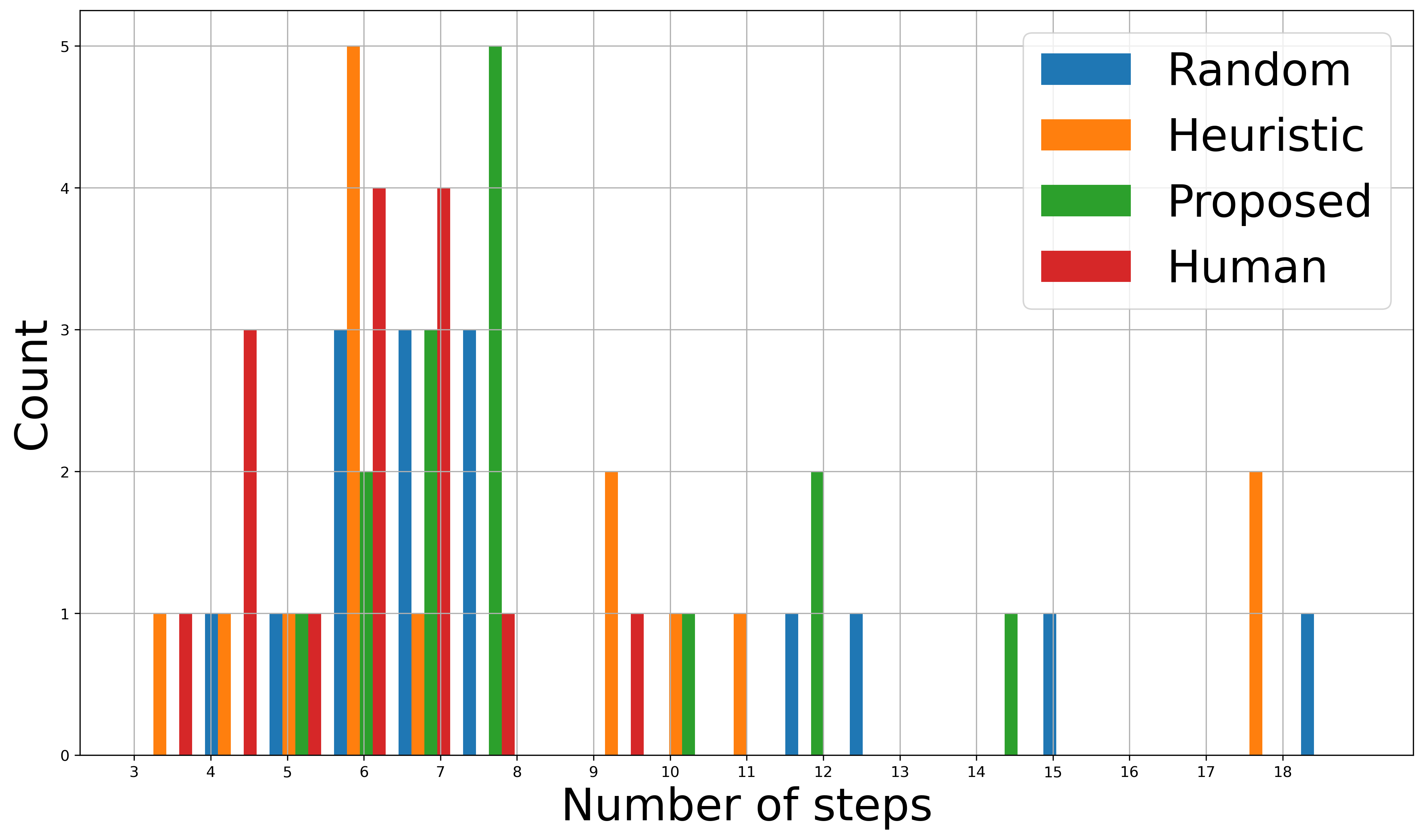}
    \caption{The occurrence of required steps of different method}
    \label{fig:hist}
\end{figure}

Fig.~\ref{fig:hist} presents a histogram of the number of steps required by our Proposed method and baselines to complete the flattening task. It shows the frequency of occurrence of different number of steps. The Human method requires at most $9$ steps to finish a task, while the Random and Heuristic methods generally need more steps than our Proposed method. Although Random and Heuristic methods may complete a task in fewer steps in some cases, on average, our method requires fewer steps than these baselines and only slightly more steps than the Human method.

\begin{table}[t]
    \caption{Number of tasks and averages steps of different methods in the experiment.}
    \label{tab:exp}
    \begin{center}
    \begin{tabular}{|c|cc|cc|}
    \hline
    \multirow{2}{*}{Method} & \multicolumn{2}{c|}{\begin{tabular}[c]{@{}c@{}}Number\\ of tasks\end{tabular}} & \multicolumn{2}{c|}{\begin{tabular}[c]{@{}c@{}}Average number \\ of steps\end{tabular}} \\ \cline{2-5} 
     & \multicolumn{1}{c|}{Easy} & Hard & \multicolumn{1}{c|}{Easy} & Hard \\ \hline
    Random & \multicolumn{1}{c|}{4} & 11 & \multicolumn{1}{c|}{7.00} & 9.36 \\ \hline
    Heuristic & \multicolumn{1}{c|}{5} & 10 & \multicolumn{1}{c|}{5.80} & 9.50 \\ \hline
    Proposed & \multicolumn{1}{c|}{3} & 12 & \multicolumn{1}{c|}{7.00} & 8.75 \\ \hline
    Human & \multicolumn{1}{c|}{4} & 11 & \multicolumn{1}{c|}{4.50} & 6.45 \\ \hline
    \end{tabular}
    \end{center}
\end{table}

\vspace{-3mm}
\section{Conclusion and Outlook}
In this paper, we proposed a promising image-processing algorithm for cloth flattening tasks that successfully infers the wrinkle distribution on the cloth and extracts a low-dimensional representation of the wrinkles. The extracted features are then used to derive a control strategy, including a stretching direction and a corresponding operation point, for robot execution. We validated the image-processing algorithm in SoftGym, and showed that our framework outperforms two other baseline methods for wrinkle removal of crumpled cloth on a table in a real robot experiment. Although the proposed method did not exceed the performance of human operators, our results demonstrate the promise of our approach for addressing this task.

In the future, we plan to incorporate learning-based methods into our framework to improve its performance. While our current algorithm outperforms two baseline methods, it still falls short of human operators' decision-making abilities. Therefore, we plan to explore Learning from Demonstration (LfD) methods to close this performance gap and enhance the quality of the control strategy generated by the algorithm. Additionally, we also want to employ this method on bi-manual robots that can prevent the cloth from slipping to get a better performance on manipulation.

\paragraph{Acknowledgements.} This work was supported by Honda Research Institute Europe GmbH as part of the project ``Learning Physical Human-Robot Cooperation Tasks'' and has been partially funded by the European Research Council Starting Grant TERI ``Teaching Robots Interactively'', project reference \#804907.
\vspace{-3mm}
%

%
%

\bibliographystyle{spmpsci}
\bibliography{root}

\begin{thebibliography}{10}
\providecommand{\url}[1]{{#1}}
\providecommand{\urlprefix}{URL }
\expandafter\ifx\csname urlstyle\endcsname\relax
  \providecommand{\doi}[1]{DOI~\discretionary{}{}{}#1}\else
  \providecommand{\doi}{DOI~\discretionary{}{}{}\begingroup
  \urlstyle{rm}\Url}\fi

\bibitem{Bersch2011}
Bersch, C., Pitzer, B., Kammel, S.: Bimanual robotic cloth manipulation for
  laundry folding.
\newblock In: IEEE International Conference on Intelligent Robots and Systems
  (IROS), pp. 1413--1419 (2011)

\bibitem{opencv_library}
Bradski, G.: {The OpenCV Library}.
\newblock Dr. Dobb's Journal of Software Tools  (2000)

\bibitem{calinon2016tutorial}
Calinon, S.: A tutorial on task-parameterized movement learning and retrieval.
\newblock Intelligent Service Robotics \textbf{9}, 1--29 (2016)

\bibitem{Caporali2020}
Caporali, A., Palli, G.: Pointcloud-based identification of optimal grasping
  poses for cloth-like deformable objects.
\newblock In: {IEEE} International Conference on Emerging Technologies and
  Factory Automation ({ETFA}), pp. 581--586 (2020)

\bibitem{Chang2020}
Chang, P., Pad{\i}r, T.: Model-based manipulation of linear flexible objects:
  Task automation in simulation and real world.
\newblock Machines \textbf{8}(3), 46 (2020)

\bibitem{Franzese2021}
Franzese, G., Mészáros, A., Peternel, L., Kober, J.: {ILoSA}: Interactive
  learning of stiffness and attractors.
\newblock In: IEEE International Conference on Intelligent Robots and Systems
  (IROS), pp. 7778--7785 (2021)

\bibitem{Hoque2020}
Hoque, R., Seita, D., Balakrishna, A., Ganapathi, A., Tanwani, A., Jamali, N.,
  Yamane, K., Iba, S., Goldberg, K.: Visuospatial foresight for multi-step,
  multi-task fabric manipulation.
\newblock In: Robotics: Science and Systems (RSS) (2020)

\bibitem{Jangir2019}
Jangir, R., Alenya, G., Torras, C.: Dynamic cloth manipulation with deep
  reinforcement learning.
\newblock In: IEEE International Conference on Robotics and Automation (ICRA),
  pp. 4630--4636 (2020)

\bibitem{Jia2017}
Jia, B., Hu, Z., Pan, J., Manocha, D.: Manipulating highly deformable materials
  using a visual feedback dictionary.
\newblock In: IEEE International Conference on Robotics and Automation (ICRA),
  pp. 239--246 (2018)

\bibitem{Joshi2017}
Joshi, R.P., Koganti, N., Shibata, T.: Robotic cloth manipulation for clothing
  assistance task using dynamic movement primitives.
\newblock In: Advances in Robotics (2017)

\bibitem{Lee1996}
Lee, T.S.: Image representation using 2d {Gabor} wavelets.
\newblock {IEEE} Transactions on Pattern Analysis and Machine Intelligence
  \textbf{18}(10), 959--971 (1996)

\bibitem{Lin2020}
Lin, X., Wang, Y., Olkin, J., Held, D.: {SoftGym}: Benchmarking deep
  reinforcement learning for deformable object manipulation.
\newblock In: Conference on Robot Learning, pp. 432--448 (2021)

\bibitem{Ma2022}
Ma, X., Hsu, D., Lee, W.S.: Learning latent graph dynamics for visual
  manipulation of deformable objects.
\newblock In: IEEE International Conference on Robotics and Automation (ICRA),
  pp. 8266--8273 (2022)

\bibitem{Marchand2005}
Marchand, E., Spindler, F., Chaumette, F.: {ViSP} for visual servoing: a
  generic software platform with a wide class of robot control skills.
\newblock IEEE Robotics and Automation Magazine \textbf{12}(4), 40--52 (2005)

\bibitem{Matas2018}
Matas, J., James, S., Davison, A.J.: Sim-to-real reinforcement learning for
  deformable object manipulation.
\newblock In: Conference on Robot Learning, pp. 734--743 (2018)

\bibitem{Miller2011}
Miller, S., Van Den~Berg, J., Fritz, M., Darrell, T., Goldberg, K., Abbeel, P.:
  A geometric approach to robotic laundry folding.
\newblock The International Journal of Robotics Research (IJRR) \textbf{31}(2),
  249--267 (2011)

\bibitem{Pinosky2022}
Pinosky, A., Abraham, I., Broad, A., Argall, B., Murphey, T.D.: Hybrid control
  for combining model-based and model-free reinforcement learning.
\newblock The International Journal of Robotics Research (IJRR) \textbf{42}(6),
  337--355 (2022)

\bibitem{Ramisa2012}
Ramisa, A., Alenya, G., Moreno-Noguer, F., Torras, C.: Using depth and
  appearance features for informed robot grasping of highly wrinkled clothes.
\newblock In: IEEE International Conference on Robotics and Automation (ICRA),
  pp. 1703--1708 (2012)

\bibitem{Salhotra2022}
Salhotra, G., Liu, I.C.A., Dominguez-Kuhne, M., Sukhatme, G.S.: Learning
  deformable object manipulation from expert demonstrations.
\newblock {IEEE} Robotics and Automation Letters \textbf{7}(4), 8775--8782
  (2022)

\bibitem{Saveriano2021}
Saveriano, M., Abu-Dakka, F.J., Kramberger, A., Peternel, L.: Dynamic movement
  primitives in robotics: A tutorial survey.
\newblock The International Journal of Robotics Research p. 02783649231201196
  (2021)

\bibitem{Singh2019}
Singh, A., Yang, L., Hartikainen, K., Finn, C., Levine, S.: End-to-end robotic
  reinforcement learning without reward engineering.
\newblock arXiv preprint arXiv:1904.07854  (2019)

\bibitem{Stria2014}
Stria, J., Průša, D., Hlaváč, V., Wagner, L., Petrík, V., Krsek, P.,
  Smutný, V.: Garment perception and its folding using a dual-arm robot.
\newblock In: IEEE International Conference on Intelligent Robots and Systems
  (IROS), pp. 61--67 (2014)

\bibitem{Sun2016}
Sun, L., Camarasa, G.A., Khan, A., Rogers, S., Siebert, P.: A precise method
  for cloth configuration parsing applied to single-arm flattening.
\newblock International Journal of Advanced Robotic Systems \textbf{13}(2), 70
  (2016)

\bibitem{Tsurumine2019}
Tsurumine, Y., Cui, Y., Uchibe, E., Matsubara, T.: Deep reinforcement learning
  with smooth policy update: Application to robotic cloth manipulation.
\newblock Robotics and Autonomous Systems \textbf{112}, 72--83 (2019)

\bibitem{Wu2019}
Wu, Y., Yan, W., Kurutach, T., Pinto, L., Abbeel, P.: Learning to manipulate
  deformable objects without demonstrations.
\newblock arXiv preprint arXiv:1910.13439  (2019)

\bibitem{Yamazaki2009}
Yamazaki, K., Inaba, M.: A cloth detection method based on image wrinkle
  feature for daily assistive robots.
\newblock In: International Conference on Machine Vision Applications, pp.
  366--369 (2009)

\bibitem{Yan2020}
Yan, W., Vangipuram, A., Abbeel, P., Pinto, L.: Learning predictive
  representations for deformable objects using contrastive estimation.
\newblock In: Conference on Robot Learning, pp. 564--574 (2021)

\bibitem{Zhang2021}
Zhang, B., Liu, P.: Model-based and model-free robot control: A review.
\newblock In: Lecture Notes in Mechanical Engineering, pp. 45--55. Springer
  Singapore (2021)

\bibitem{Zhu2022}
Zhu, J., Cherubini, A., Dune, C., Navarro-Alarcon, D., Alambeigi, F., Berenson,
  D., Ficuciello, F., Harada, K., Kober, J., Li, X., Pan, J., Yuan, W.,
  Gienger, M.: Challenges and outlook in robotic manipulation of deformable
  objects.
\newblock {IEEE} Robotics \& Automation Magazine \textbf{29}(3), 67--77 (2022)

\end{thebibliography}

\end{document}